\pdfoutput=1
\documentclass[11pt]{article}

\usepackage[final]{acl}

\usepackage{times}
\usepackage{latexsym}
\usepackage{multirow}
\usepackage{amsmath}
\usepackage[T1]{fontenc}

\usepackage[utf8]{inputenc}

\usepackage{microtype}

\usepackage{inconsolata}

\usepackage{graphicx}

\title{Dependency Parsing with the Structuralized Prompt Template}

\author{Keunha Kim \\
  SungKyunKwan University \\
  Republic of Korea \\
  \texttt{keunhakim98@gmail.com} \\\And
  Youngjoong Ko\thanks{Corresponding author} \\
  SungKyunKwan University \\
  Republic of Korea \\
  \texttt{yjko@skku.edu} \\}

\begin{document}
\maketitle
\begin{abstract}
Dependency parsing is a fundamental task in natural language processing (NLP), aiming to identify syntactic dependencies and construct a syntactic tree for a given sentence. Traditional dependency parsing models typically construct embeddings and utilize additional layers for prediction. We propose a novel dependency parsing method that relies solely on an encoder model with a text-to-text training approach. To facilitate this, we introduce a structured prompt template that effectively captures the structural information of dependency trees. Our experimental results demonstrate that the proposed method achieves outstanding performance compared to traditional models, despite relying solely on a pre-trained model. Furthermore, this method is highly adaptable to various pre-trained models across different target languages and training environments, allowing easy integration of task-specific features.
\end{abstract}
\section{Introduction}

\begin{figure}[h]
 \includegraphics[width=\columnwidth]{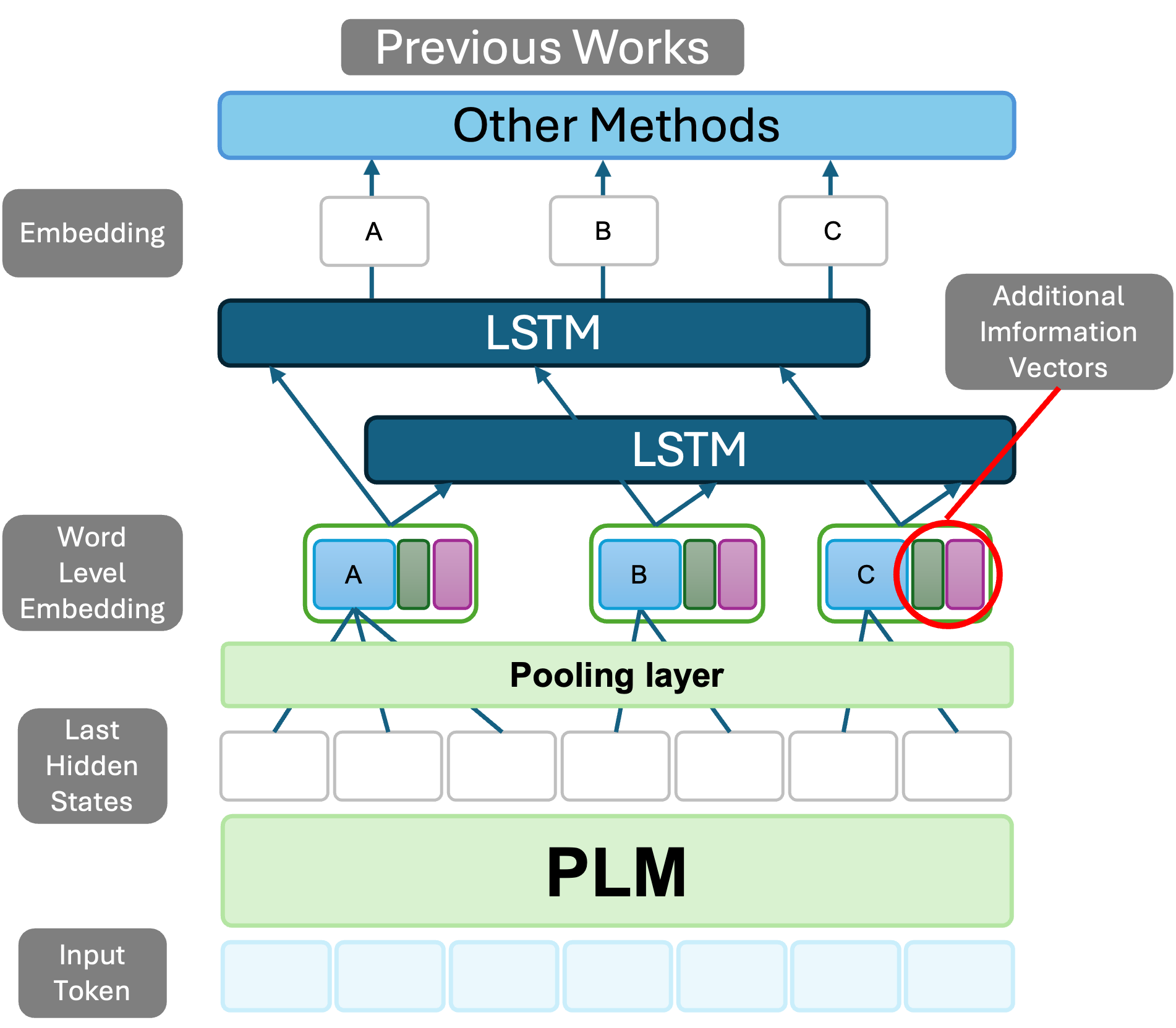}
 \caption{A figure shows that pipeline of traditional dependency parsing}
 \label{fig:traditional}
\vskip -.24in
\end{figure}

Dependency parsing is a fundamental task in natural language processing (NLP) that analyzes syntactic relationships between words in a sentence. Figure \ref{fig:traditional} illustrates the traditional dependency parsing pipeline. Traditionally, dependency parsing is performed in two steps: (1) creating word-level embeddings, and (2) identifying the head word of each word along with its dependency relation using the generated embeddings.

Previously, dependency parsing primarily relied on simple pre-processed contextual vectors to initialize embeddings \cite{li-etal-2018-seq2seq,strzyz-etal-2019-viable,vacareanu-etal-2020-parsing}. With the advent of powerful pre-trained language models such as BERT \cite{bert}, recent dependency parsing approaches leverage these models to initialize word embeddings, achieving superior performance compared to earlier methods \cite{amini-etal-2023-hexatagging}. In the second step of dependency parsing, previous studies have shown that graph-based methods, such as biaffine \cite{dozat-biaffine}, yield good performance in identifying relations. Consequently, this approach was extended to learn the subtree information of the dependency tree \cite{yang-tu-2022-headed}. Since the structural characteristics of dependency trees increased training complexity and difficulties, some studies use the sequence tagging method for parsing \cite{li-etal-2018-seq2seq,amini-cotterell-2022-parsing}. These approaches add simple layers after embedding construction and label the words into structural information sequentially. In particular, the hexatagging method achieved state-of-the-art performance by generating structural information with a finite set of tags through decoding \cite{amini-etal-2023-hexatagging}. 
In addition, \citet{lin2022dependency} demonstrated that encoder-decoder models effectively generate relation unit texts from input texts, highlighting that dependency parsing can be performed solely using pre-trained language models. This study is particularly significant as it explores dependency parsing by transforming modified text input into structured dependency output. However, the notable limitation of this approach lies in the increased computational time required for parsing, as the number of tokens grows due to the generation-based method.

In this paper, we propose a novel method to perform dependency parsing solely on pre-trained encoder models that are constructed by prompt engineering using additional tokens as soft prompts. We hypothesize that prompt engineering can effectively convert the text-to-structure task in dependency parsing to the text-to-text task by pre-trained language models, just like the sequence labeling method. Hence, the output text sequence of the proposed method has to reflect the tree structure of dependency parsing well. To achieve this, we design several soft prompts so that our model can identify the structural information of the tree structure, and then apply the \textbf{S}tructuralized \textbf{P}rompt \textbf{T}emplate (SPT) for each processing unit of dependency parsing using the developed soft prompt. We believe that prompt learning with the structuralized prompt template enables effective and efficient dependency parsing only on the pre-trained language models. Eventually, by learning through the structuralized prompt template, the \textbf{S}tructuralized \textbf{P}rompt \textbf{T}emplate based \textbf{D}ependency \textbf{P}arsing (SPT-DP) method achieves the high performance and simplified training by reducing the gap between pre-training and fine-tuning because it is based on only the pre-trained language models for the text-to-text task. 
As a result, the performance of the proposed method surpasses most of existing methods: 96.95 (UAS) and 95.89 (LAS) on English Penn Treebank (PTB; \citet{marcus-etal-1993-building}). On the 2.2 version of Universal Dependencies (UD 2.2; \citet{UD2.2}), it obtains the state-of-art performance in 2 languages out of 12 languages when using the cross-lingual Roberta model \cite{roberta}. Since our method utilizes only a single encoder model, our approach is faster than existing methods. In particular, it achieves a 40x speed improvement compared to DPSG, which employs a similar text-to-text approach. Furthermore, our method achieves a performance comparable to that of the SOTA model with a complicated and heavy architecture in the Korean Sejong dataset.

\section{Preliminaries}

\subsection{Dependency Parsing}
Dependency parsing is the process of identifying dependency relationships among words in a sentence. Given a sentence $S = (w_1, w_2, \ldots, w_n)$, the task is to derive the dependency relations, $R_S = (r_1, r_2, \ldots, r_n)$, for each word. Each relation, $r_i = (H_i, L_{(w_i,w_{H_i})})$, consists of two components: (1) the index of the head word $H_i$ and (2) the dependency relation label $ L_{(w_i,w_{H_i})}$, between $w_i$ and $w_{H_i}$, where $ L_{(w_i,w_{H_i})} \in \mathcal{L}$ and $\mathcal{L}$ is the set of predefined dependency relation labels. This study focuses on syntactic dependency parsing that analyzes grammatical dependency relations within sentences. Depending on the definition of the syntactic dependency tree, each word has exactly one parent node, ensuring a hierarchical structure. Therefore, the ultimate goal of dependency parsing is to analyze the sentence $S$ into the form $S_{dep} = \{(w_i, r_i) \mid r_i = (H_i, L_{(w_i,w_{H_i})}), i \in \{1, 2, \ldots, n\}\}$, where each word $w_i$ is paired with its corresponding dependency relation $r_i$. \footnote{ Table \ref{tab:notation} for a list of notations along with explanations.}

\begin{figure}[h]
    \begin{center}
        $S = (He, loves, his, rabbits)$
    \end{center}
    \includegraphics[width=\columnwidth]{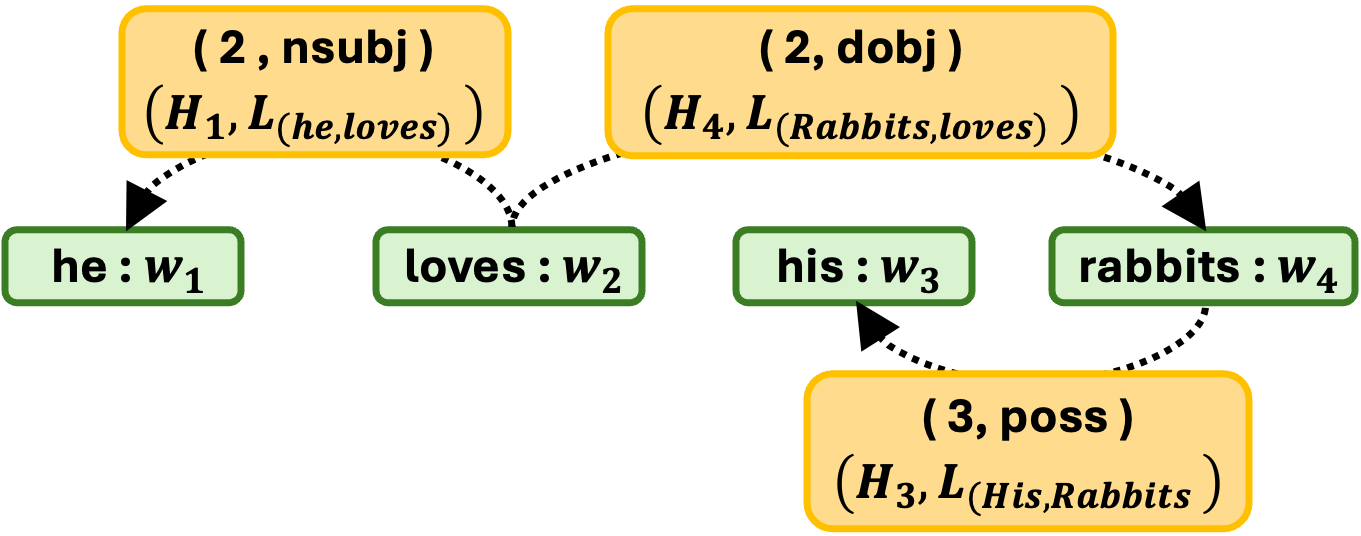}
    \centering
    \resizebox{\columnwidth}{!}{
    \begin{tabular}{c|c|c|c}
        \hline
        \hline
        \textbf{Word} : \textbf{$w_i$} & \textbf{Index} : \textbf{ $i$} & \textbf{Head} : \textbf{ $H_i$} & \textbf{Label} : \textbf{$ L_{(w_i,w_{H_i})}$}\\        
        \hline
        \hline
        he      & 1  & 2  & \texttt{[nsubj]}  \\
\hline
        loves   & 2  & 0(root)  & \texttt{[root]}   \\
      
\hline  his     & 3  & 4  & \texttt{[poss]}   \\
      \hline  rabbits & 4  & 2  & \texttt{[dobj]}   \\
      \hline 
    \end{tabular}}
\caption{Dependency Parsing Example}
\vskip -.12in
\end{figure}

\subsection{Prompt based tuning}
Initially, the concept of a prompt emerged as a method for training models, particularly to bridge the gap between the pre-training and fine-tuning stages. PET (Pattern-Exploiting Training) proposed a method to reduce the gap between pre-training and fine-tuning by using cloze-style patterns instead of the traditional [CLS] token-based classification approach in BERT \cite{schick-schutze-2021-exploiting}. This allows the pre-trained language model to naturally adapt to classification tasks. For example, in sentiment analysis, a sentence like "This movie is really [MASK]." is used, and the predicted probability of the [MASK] token is used to determine its sentiment.

Subsequently, prompts have been explored in combination with methods like PEFT (Parameter-Efficient Fine-Tuning) to improve model efficiency and task performance \cite{li-liang-2021-prefix, lester-etal-2021-power}. In generative models like GPT-3 \cite{NEURIPS2020_1457c0d6}, text-based prompts are used to guide and shape the outputs into task-specific outputs.

In this study, prompts are used to provide additional linguistic hints, such as grammatical structures or contextual information, necessary for dependency parsing. Specifically, prompts are added to the encoder model through preprocessing. This is done by concatenating prompt phrases, such as task-specific instructions, to the original input sentence. Prompts are applied to each word in the input sentence as shown in the following formulation:
\begin{equation}
P(S) = (T(w_1), T(w_2), \ldots, T(w_n))
\end{equation}

where $P(S)$ represents the transformed sentence and $T(w_i)$ is the prompt template applied to the $i$-th word $w_i$.

\subsection{Pre-trained Language Model}

Transformer-based pre-trained models are built upon the following architectures:

\begin{itemize}  
\item \textbf{Encoder-based Models}: These models encode input text to perform tasks like masked token prediction or classification by adding a linear layer. They are also widely used in \textit{Dense Passage Retrieval (DPR)} \cite{karpukhin-etal-2020-dense} to extract representative embeddings for documents. $X \rightarrow X’$
  \item \textbf{Decoder-based Models}: These models take an initial token or prompt and generate tokens sequentially through autoregressive decoding. $X \rightarrow Y$
\end{itemize}
The traditional approach to dependency parsing involves extracting token-level embeddings from encoder-based pre-trained models, integrating them at the word level, and constructing word embeddings. These embeddings are then analyzed through various methods to perform dependency parsing. In this process, the performance of dependency parsing heavily depends on how effectively the encoder provides information relevant to syntactic structures. 

\begin{figure*}[h]
 \includegraphics[width=1\linewidth]{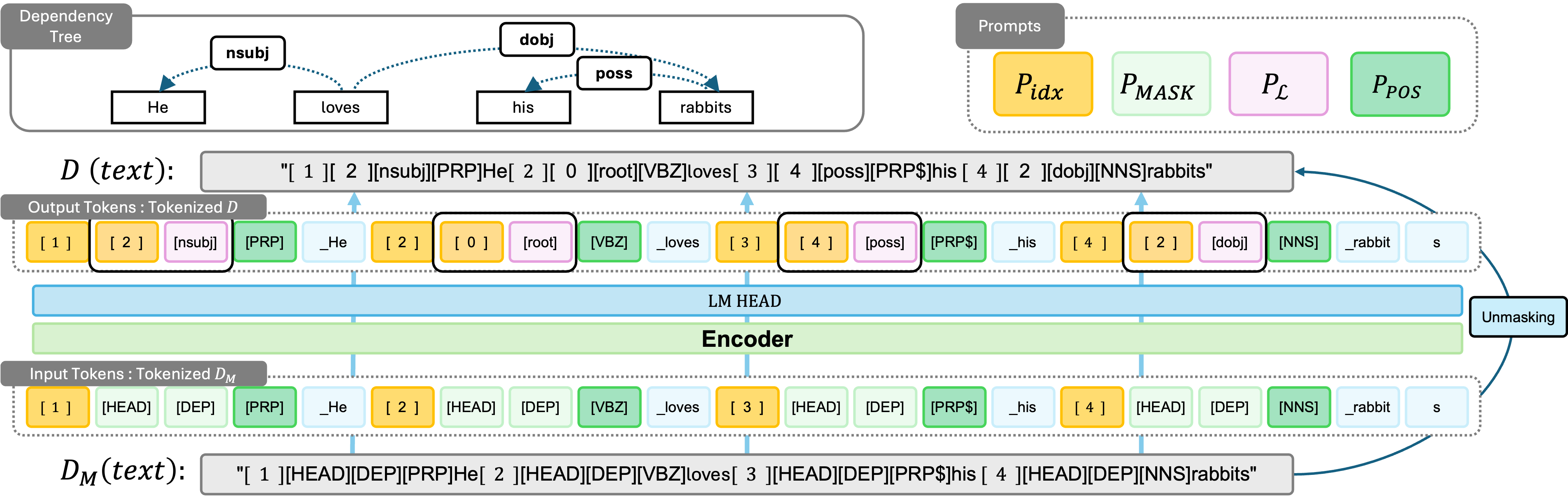}
 \caption {Overview of the Structuralized Prompt Template based Dependency Parsing (SPT-DP) method.}
  \label{fig:main}
\end{figure*}

\section{Structuralized Prompt Template}
As aforementioned, traditional methods approach dependency parsing by utilizing predefined embeddings processed through specific modules. However, Dependency Parsing via Sequence Generation (DPSG) departed from this paradigm and introduced a text-to-text dependency parsing method \cite{lin2022dependency}. Despite its innovative nature, DPSG faced limitations due to its generation-based approach, which could produce unintended outputs. In addition, the need to generate tokens one by one resulted in slower processing speeds, especially when processing additional tokens.
To overcome these challenges, we propose a novel approach that leverages encoder models. Inspired by the ability of encoders to facilitate text-to-text learning through pre-training tasks such as Masked Language Modeling (MLM), we devised a method to represent dependency parsing information as text and apply it during training. We call this method \textbf{S}tructuralized \textbf{P}rompt \textbf{T}emplate (SPT). Unlike traditional dependency parsing, which treats text and dependency structures as separate levels, SPT integrates dependency information directly into text sequences. This approach enables dependency parsing entirely within the textual level by eliminating additional integration steps and improving overall efficiency.

\subsection{Dependency Parsing with Text Representation}

Basically, dependency parsing is the task of finding a word with a dependency relation and determining its corresponding dependency relation label for each word. To express the structured dependency relationships through the prompt template for each word, several conditions should be satisfied: 1) each template must be distinguished by its pattern through whole training, 2) it must be able to indicate its index, 3) it must be able to refer to the other word template with dependency relationship through the output, and 4) it must be able to express the dependency relation label through the output. 

In the first condition, we ensure that the language model can distinctly recognize each template by following a consistent pattern rather than a specific token through all the training process. To satisfy the second and third conditions, we add index prompts $P_{idx}$ that serve two roles: representing the template and indicating the referred template’s index. 
Because the $P_{idx}$ prompts represent the index of templates in a formatted text sequence, each template can refer to another template within a dependency relation regardless of the input sequence using the $P_{idx}$ prompts. Finally, the fourth condition is resolved by adding the dependency relation labels as prompt $P_{\mathcal{L}}$. In addition, we add $P_{POS}$ to reflect information for part-of-speech (POS) analysis. However, several challenges arise when using this prompt approach for dependency parsing:
\begin{enumerate}
\item{\textbf{Semantic Representation Issues}: If the prompts are directly integrated into the model’s existing vocabulary without introducing new tokens, it may disrupt the original vocabulary structure, making it difficult to capture meaningful semantics.}
\item{\textbf{Variability in Length}: The explicit addition of prompts, such as numerical indices and dependency relation labels, causes variations in the length of the enhanced sentence. Masking these components can lead to further inconsistencies in sequence length, which is critical for learning in encoder models.}
\end{enumerate}

To address these issues, we propose to add each prompt as a new token to the vocabulary. 
The above prompts are individually added to the vocabulary enclosed in []. This action allows them to be added independently to the existing vocabulary without any duplications. 
This approach allows the model’s training process to be examined at the text level.
To avoid confusion, we clarify that the notations $P_{idx}$ , $P_{\mathcal{L}}$ , $P_{POS}$ , and $P_{MASK}$\footnote{to be explained in Section \ref{sec:mask}.} represent string text encapsulated in square brackets [].
\begin{equation}
T(w_i) = \text{"}\overbrace{\underbrace{[\,i\,]}_{P_{abs}}\underbrace{[H_i]}_{P_{ref}}}^{P_{idx}}\overbrace{[L_{(w_i,w_{H_i})}]}^{P_{\mathcal{L}}}\overbrace{[POS_{w_i}]}^{P_{POS}}w_i\text{"}
\label{eqa:preprocess}
\end{equation}
\begin{equation}
D = (T(w_1),T(w_2),...,T(w_n))
\label{eqa:dep}
\end{equation}

The final word-level prompt template is represented by Equation \ref{eqa:preprocess}: for each word $w_i$, four prompts are added. In the first prompt $[\,i\,]$ is $P_{abs}$, which represents the absolute index token indicating the index of each word in the sentence.
The second prompt $[H_i]$ is $P_{ref}$ that specifies the index of the word referenced by the given word $w_i$. The only distinction between $P_{abs}$ and $P_{ref}$ is their positions in the template; they serve different roles but share the same tokens. That is why they are grouped under $P_{idx}$. The third is the dependency relation label $[L_{(w_i,w_{H_i})}]$, and the fourth contains $[POS_{w_i}]$ that is POS-tag of $w_i$. The proposed \textbf{S}tructuralized \textbf{P}rompt \textbf{T}emplate (SPT) can effectively incorporate the dependency parsing information into input sentence using these prompts. That is, by applying this template to every word in the sentence, we construct a modified sentence $D$ that encapsulates dependency parsing information. From a certain perspective, performing dependency parsing can be regarded as transforming the original sentence $S$ into a Prompted sentence $D$.

\begin{figure}[h]
 \includegraphics[width=\columnwidth]{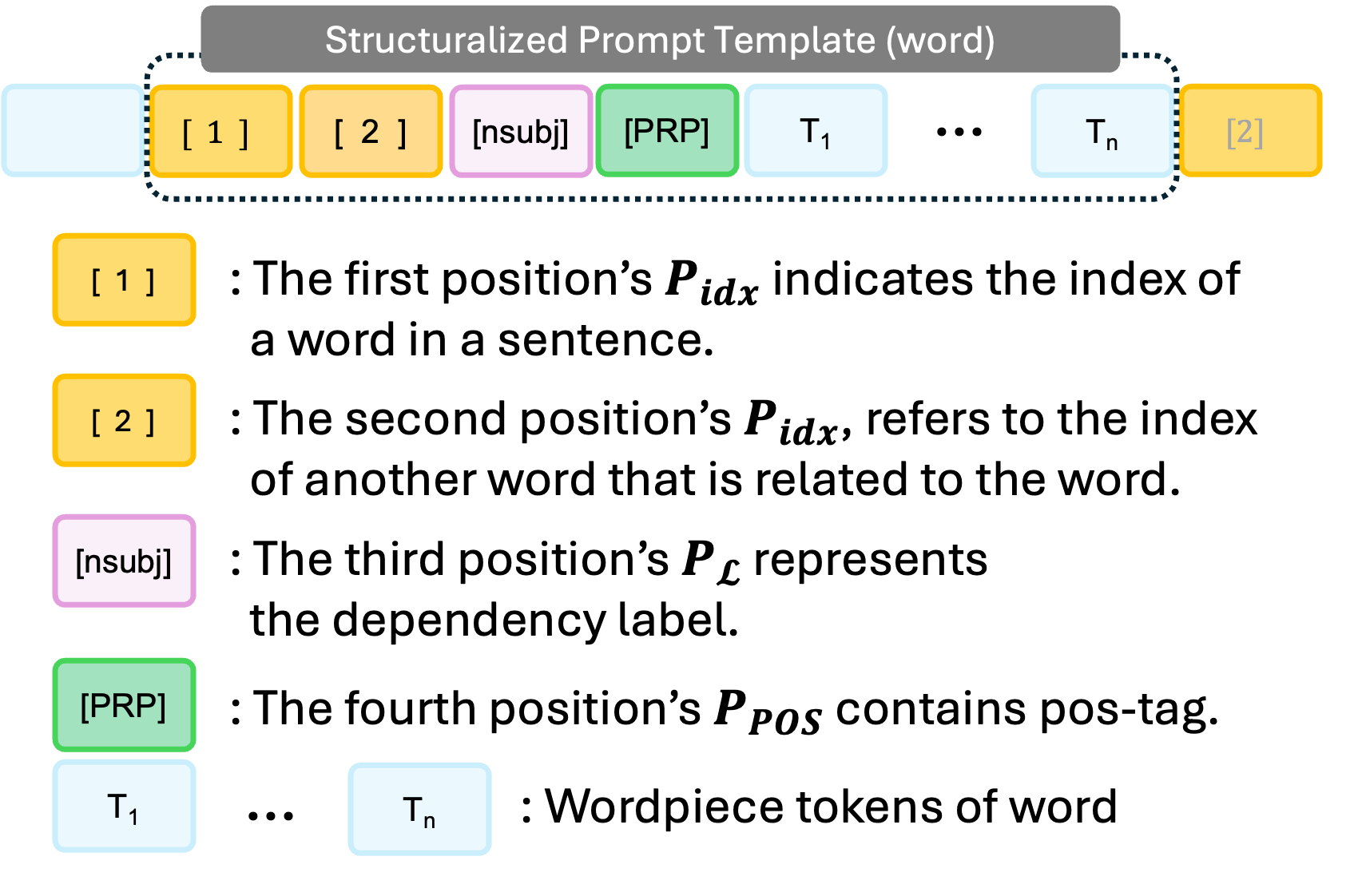}
 \caption{A figure of word level SPT example}
 \label{fig:set}
 \vskip -.12in
\end{figure}
\subsection{Prediction Task using Soft Prompts $P_{MASK}$ : [HEAD] and [DEP]}

\label{sec:mask}
For the training, $P_{MASK}$, which contains the [HEAD] and [DEP] tokens, serves as a soft prompt and plays a role similar to the masked token in the Masked Language Model (MLM) task. Since performing dependency parsing involves predicting $[H_i]$ and $[L_{(w_i,w_{H_i})}]$, the corresponding parts in the equation are masked to enable the model to learn by making predictions. In this process, $[H_i]$ is masked as [HEAD] and [$L_{(w_i,w_{H_i})}]$ is masked as [DEP], respectively.
The $P_{MASK}$ are also added in the tokenizer's vocab for prompt engineering.
\begin{equation}
M(T(w_i)) = \text{"}\overbrace{[\,i\,]}^{P_{abs}}\overbrace{[HEAD][DEP]}^{P_{MASK}}[POS_{w_i}]w_i\text{"}
\label{eqa:masking}
\end{equation}
\begin{equation}
D_M = (M(T(w_1)),M(T(w_2)),...,M(T(w_n)))
\label{eqa:masked}
\end{equation}

As aforementioned, the $P_{MASK}$ are used to infer two main prediction tasks for the head word and the dependency relation label. They are arranged in the second and third positions of the structuralized prompt templates, forming a consistent pattern in the input text sequence. In our approach, the model is fine-tuned to predict head word and dependency relation labels by the [HEAD] and [DEP] prompts.

Ultimately, the goal is to learn to reconstruct the $D$ from the masked input $D_M$, which has been structured using prompts. This involves different prediction mechanisms for encoder and decoder models:

\paragraph{Encoder-only Models} The encoder model learns by reconstructing the masked input $D_{M}$ into its original form $D$. Since the output sequence retains the same structure as the input sequence, the model effectively maps $D_{M}$ back to $D$:
$Enc(D_{M}) \rightarrow D$
\paragraph{Decoder-based Models} In encoder-decoder models, the encoder processes the masked input $D_{M}$ to generate a latent representation and the decoder uses it to reconstruct $D$. On the other hand, $D_{M}$ of decoder-only models serves as a prefix and it guides the decoder to generate the unmasked $D$, autoregressively:
$Gen(D_{M}) \rightarrow D$ 

\section{Prompt-based Training}

As mentioned earlier, in the encoder model, training is conducted in an MLM manner, where the input masked with $P_{MASK}$:[HEAD] and [DEP] are used to predict reference indices $P_{ref}$:$[H_i]$ and dependency relation label $P_{\mathcal{L}}$:$[L_{(w_i,w_{H_i})}]$ prompts. Since we have added each prompt to the tokenizer, the model is trained to output the corresponding prompt IDs through the LM head. In the decoder model, the training process involves providing masked input and learning to generate the unmasked text.

The loss function for training is calculated by the following equations. In Equation \ref{eqa:one}, $X_{input}$ is the tokenized $D_M$ that is a concatenated sequence of SPTs where $[H_i]$ and $[L_{(w_i,w_{H_i})}]$ are replaced by [HEAD] and [DEP] prompts. In Equation \ref{eqa:two}, $Y_{label}$ is the tokenized $D$ that is the prompted sequence based on SPT.

Since all the prompts are added to the tokenizer's vocabulary, the lengths of $X_{input}$ and $Y_{label}$ are always the same, which is a crucial condition in encoder-only models. In summary, the models are trained with $D_M$ as input and $D$ as the label. Encoder-only models are trained with $\mathcal{L}_{enc}$, which optimizes token prediction based on $X_{input}$ following BERT’s masked language modeling (MLM) for bidirectional contextual learning. Decoder-based models use $\mathcal{L}_{dec}$, where each token $y_i$ is generated sequentially based on $X_{input}$ and previous outputs $Y_{<i}$, following an autoregressive approach similar to GPT. $\theta$ is the parameter of model.

The objective function is based on \textbf{Cross-Entropy Loss}, defined as follows:
\begin{equation}
Tokenize(D_M) = X_{input} = [x_{1},x_{2},...,x_{N}]
\label{eqa:one}
\end{equation}
\begin{equation}
Tokenize(D) = Y_{label} = [y_{1},y_{2},...,y_{N}]
\label{eqa:two}
\end{equation}
\begin{equation}
\mathcal{L}_{enc} = -\sum_{i=1}^N \log P(y_{i} | X_{input};\theta)
\label{eqa:loss}
\end{equation}
\begin{equation}
\mathcal{L}_{dec} = -\sum_{i=1}^N \log P(y_{i} | X_{input}, Y_{<i};\theta)
\label{eqa:loss2}
\end{equation}

In summary, $\mathcal{L}_{enc}$ and $\mathcal{L}_{dec}$ correspond to Cross-Entropy Loss, where the model optimizes token prediction probabilities. In encoder-only models, $\mathcal{L}_{enc}$ applies token-wise classification in an MLM setting, while in decoder-based models, $\mathcal{L}_{dec}$ follows an autoregressive generation paradigm.

\begin{table*}
 \setlength{\abovecaptionskip}{-8pt}
	\centering
	{ 
		\makebox[\linewidth]{\resizebox{\linewidth}{!}{%
    \begin{tabular}{lccccccccccccc}\hline
            & 	bg   & 	 ca   &  cs   &  de   &  en  &  es  &  fr   &  it  & nl  & no  & ro  & ru   &  Avg. \\ \hline\hline
            \citet{dozat-biaffine}$\diamondsuit$   &  90.30   &  \textbf{94.49}   &  {92.65}   &  \textbf{85.98}   &  91.13   &  \underline{93.78}   &  \textbf{91.77}   & \underline{94.72}   & 91.04   & 94.21   & 87.24   &  94.53   & 91.82 \\ 
        \citet{wang-tu-2020-second}$\diamondsuit$   &  91.30   &  93.60   &  92.09   &  82.00   &  90.75   &  92.62   &  89.32   &  93.66   &  91.21   &  91.74   &  86.40   &  92.61    &  90.61 \\ 
         \citet{yang-tu-2022-headed}$\diamondsuit$   &  91.10   &  \underline{94.46}   & 92.57   &  \underline{85.87}  &  \underline{91.32}   &  \textbf{93.84}   &  \underline{91.69}   &  \textbf{94.78}   &  {91.65}   &  \underline{94.28}   &  {87.48}   & 94.45   &  \underline{91.96} \\  
         \citet{lin2022dependency}*   &  \textbf{93.92}   &  93.75   & 92.97   &  84.84   &  \textbf{91.49}   &  {92.37}   &  90.73   &  {94.59}   &  \underline{92.03}   & \textbf{95.30}   &  88.76   & \textbf{95.25}   &  \textbf{92.17} \\  
         \citet{amini-etal-2023-hexatagging}$\diamondsuit$ &  92.87  &  93.79   &  92.82 &  85.18   &  90.85   &  93.17   &  91.50   &  \underline{94.72}   &  91.89   &  {93.95}   &  87.54   &  94.03   &  {91.86}
         \\ \hline			
         SPT-DP (multilingual BERT)   &  91.20  &  90.81   &  92.22   &  79.68   &  87.36   &  90.33   &  88.31   &  92.00   &  89.37   &  90.64   &  86.12   &  93.17   &  89.27  \\
         SPT-DP (XLNet-large)   &  -  &  -   &  -   &  -   &  90.58   &  -   &  -   &  -  &  -  &  -   &  -   &  -   &  - \\
         SPT-DP (XLM-RoBERTa-large) & 93.11  &  92.54   &  \textbf{94.14}   &  82.11   &  88.50   &  91.69   &  88.02   &  93.16   &  91.15   &  93.13   &  \textbf{88.87}   & \underline{95.12}  & 90.90   \\
         SPT-DP (global) & \underline{93.91} & 92.85 & \underline{93.27} & 82.91 & 89.27 & 92.03 & 89.96 & 93.24 & \textbf{92.28} & 92.81 & \underline{88.84} & 94.39 & 91.31  \\
         \hline \\
    \end{tabular}}}
 }
	\caption{12 languages' LAS scores on the test sets in UD 2.2. $\diamondsuit$ use multilingual BERT for embedding and * uses T5-base model for sequence generation parsing. Best and second-best scores are in \textbf{bold} and \underline{underlined}}
	\label{tab:ud} 
\end{table*}

\begin{table}
\centering
 \setlength{\abovecaptionskip}{4pt}
\resizebox{\columnwidth}{!}{
\begin{tabular}{@{}lcc@{}}\hline
  & \multicolumn{2}{c}{PTB} \\
Model  & UAS & LAS \\ \hline\hline
\citet{zhou-zhao-2019-head}*        & 97.0  & 95.43 \\
\citet{mrini-etal-2020-rethinking}*  & 97.42  & 96.26 \\ \hline
\citet{dozat-biaffine}     \;\;\;  \;\;          & 95.74  & 94.08  \\
\citet{wang-tu-2020-second}          & 96.91  & 95.34 \\
\citet{yang-tu-2022-headed}      & \underline{97.24}  & 95.73   \\
\citet{lin2022dependency}      & 96.64  & 95.82  \\
\citet{amini-etal-2023-hexatagging}     & \textbf{97.4}  & \textbf{96.4} \\ \hline
SPT-DP (XLNet-large)             & 96.95  & \underline{95.88}\\
 \hline
\end{tabular}}
\caption{Results on PTB dataset. * use additional constituency parsing information so they are not comparable to other methods.}
\label{tab:ptb_results}

\end{table}

\section{Experiments}
We first apply our approach to two datasets used in previous studies, including \textbf{PTB} (Penn Treebank) and \textbf{UD 2.2} \cite{UD2.2} covering 12 languages. Since these datasets primarily consist of \textbf{fusional} languages in which a single morpheme can encode multiple grammatical features, we extended our experiments to \textbf{agglutinative} languages. Unlike fusional languages, agglutinative languages express grammatical relationships through sequences of distinct morphemes, which are carrying out their individual function. For this, we select the \textbf{Sejong} corpus, a Korean language dataset, as an agglutinative language dataset.
\subsection{Datasets}

\paragraph{PTB } This English data is preprocessed by Stanford Parser v3.3.0 \cite{de2008stanford} to convert it into CoNLL format, following the approach of \citet{mrini-etal-2020-rethinking}. 
\paragraph{UD 2.2}
\vskip -.12in
This is composed of 12 languages from UD dataset v2.2 and we follow previous work \cite{amini-etal-2023-hexatagging} for data splitting and organizing. The POS tag information is not used for the experiments by omitting $P_{POS}$ in the template.\vskip -.10in
\paragraph{Sejong}
\vskip -.12in 
This is the Korean dataset and only the POS tags of the first and last morphemes are used for this experiment because Korean words consist of multiple morphemes.

\subsection{Pre-trained Language Models}  

\paragraph{Encoder-based Models}  
XLNet-large, Multilingual BERT, XLM-RoBERTa-large, and RoBERTa
\paragraph{Decoder-based Models}  
\vskip -.10in
T5-base, BART-large, LLaMA3.2-3B, and Qwen2.5-3B  

\subsection{Comparison Models}
\citet{zhou-zhao-2019-head} and \citet{mrini-etal-2020-rethinking} used additional constituency parsing information so they are not comparable to other methods directly.
\citet{dozat-biaffine} introduced a biaffine model as a graph-based dependency parsing approach. \citet{wang-tu-2020-second}  proposed a second-order graph-based method with message passing. \citet{yang-tu-2022-headed} developed a projective parsing method based on headed spans. \citet{lin2022dependency} introduced a sequence generation-based parsing method, while \citet{amini-etal-2023-hexatagging} leveraged structural tags and sequential tag decoding. \citet{park-2019} and \citet{IM-2021} constructed a dependency parser using the Korean morpheme version of BERT. 

\begin{table}[h]
\centering
\resizebox{\columnwidth}{!}{
\begin{tabular}{@{}lcc@{}}\hline
Model  & UAS & LAS \\ \hline\hline
 \citet{park-2019} \;  & 94.06  & 92.00 \\ 
 \citet{IM-2021} \; \; \; \; \; \; \;   & \textbf{94.76}  & \textbf{92.79} \\ \hline
SPT-DP  & \underline{94.52}   & \underline{92.36}\\
 \hline
\end{tabular}}
\caption{Results on Sejong dataset}
\label{tab:sejong}

\end{table}

\subsection{Experimental Results}

\begin{table*}[h]
 \setlength{\abovecaptionskip}{-4pt}
	\centering
	{ 
		\makebox[\linewidth]{\resizebox{\linewidth}{!}{%
 \begin{tabular}{c|cccccccccccc|c}\hline
  & 	bg& 	 ca&  cs&  de&  en  &  es  &  fr&  it  & nl  & no  & ro  & ru&  Avg. \\ \hline\hline
base & 93.11  &  92.54   &  \textbf{94.14}   &  82.11   &  88.50   &  91.69   &  88.02   &  93.16   &  91.15   &  \textbf{93.13}   &  \textbf{88.87}   & \textbf{95.12}  & 90.90   \\

\hline
global & \textbf{93.91} & \textbf{92.85} & 93.27 & \textbf{82.91} & \textbf{89.27} & \textbf{92.03} & \textbf{89.96} & \textbf{93.24} & \textbf{92.28} & 92.81& 88.84 & 94.39 & \textbf{91.31} \\
unseen & 81.12 & 88.49 & 79.83 & 75.80 &  72.59 & 87.26 & 81.30 & 79.63& 78.17 & 77.86 & 77.80 & 77.05 & 79.74\\
\hline
\end{tabular}}}
 }
 \vspace{10pt}
\caption{LAS scores for the UD2.2 dataset. The first row lists the languages in the test set. \textbf{base} represents the performance of a model trained individually for each language. \textbf{global} represents the performance of a single multilingual model trained on all languages. \textbf{unseen} represents the performance of individual models trained on all languages except for the test language, evaluating zero-shot transfer performance.}
	\label{tab:global} 
\vskip -.14in
\end{table*}

Table \ref{tab:ptb_results} presents the performance comparison of different models on the PTB dataset. Our proposed method achieves competitive performance compared to state-of-the-art (SOTA) models in spite of relying \textbf{solely on a pre-trained language model} without additional complex modules. That is, despite its \textbf{lightweight and efficient design}, our method is ranked among the top-performing models.

As shown in Table \ref{tab:ud}, our method exhibits lower performance with multilingual BERT on UD 2.2. However, using a larger model, XLM-RoBERTa-large, leads to significant performance improvements across most languages and it eventually achieves SOTA results on two languages. Similarly, our approach with XLNet-large on UD 2.2 English data shows more improved performance depending on its English-only pre-training. 
We can think that the larger models with richer contextual representations we can use, the better performance our method can obtain. This is because we can effectively leverages their expressiveness to enhance parsing performance.
In addition, we construct the \textbf{global}\footnote{This approach will be discussed in detail in \ref{sec:multi}} model by training on the entire languages of UD 2.2 dataset and it leads to overall improvements in parsing accuracy across many languages. Notably, in \textbf{Dutch (nl)}, our approach achieves an additional SOTA result. On the other hand, the results on the \textbf{Sejong} dataset (Table \ref{tab:sejong}) demonstrate that our method achieves performance on par with more complex SOTA models, even though it is based on \textbf{lighter and more efficient architecture} for \textbf{agglutinative languages}.

\section{Analysis}

\subsection{Unified Cross-lingual Dependency Parsing}
\label{sec:multi}
Furthermore, our evaluation of language-specific experiments on the UD 2.2 dataset is expanded to cross-lingual experiments. Since the composition of dependency relation labels varies across languages, we integrate dependency relation labels from 12 languages into a shared vocabulary to construct a unified model for cross-lingual dependency parsing. 
Then we train a single model using a unified training dataset based on integrated dependency relation labels and we refer to it as the \textbf{global} model in Table \ref{tab:global}. This cross-lingual model exhibits robust performance and it demonstrates its ability to generalize multi-languages while maintaining competitive results. As you can see in Table 4, it outperforms the \textbf{base}\footnote{A trained model where the test language and train language are the same.} model in many languages. In addition, we evaluate a model trained by excluding the training set of the target test language, referred to as the \textbf{unseen} model, for an out-of-domain evaluation. This \textbf{unseen} model achieves meaningful performance even on languages for which it was not explicitly trained in dependency parsing. This highlights cross-lingual correlations in dependency parsing and shows that our approach provides flexibility and scalability, particularly in multilingual or resource-constrained scenarios.

\subsection{Length Robustness}
\label{sec:length}
In previous studies, dependency parsing first attempt to represent the syntactic information of words in a sentence by feeding the final hidden states from the pre-trained model into additional modules and classifies each embedding or directly compares the output embeddings of each word to find dependency relations.
In contrast, our study newly defines and utilizes index prompts, $P_{idx} $($P_{abs}$, $P_{ref}$); $P_{ref}$ in SPT of a word indirectly refer to the $P_{abs}$ in SPT of another word to represent their dependency relations. Therefore, we have to check out how the relations between $P_{idx}$ tokens are well trained by the proposed method. Table \ref{tab:length} presents the performance according to sentence length, showing a tendency for performance to decrease as sentence length increases. Furthermore, as shown in Table \ref{tab:length2}, which details performances based on the index range of templates, predictions for dependency relations in higher indices tend to have lower accuracy. 
Statistically, because longer sentences have relatively small population, prompts corresponding to higher indices may not have been sufficiently learned. In future work, we should focus on developing improved training strategies to enhance performances for high indexed words in longer sentences.

\begin{table}[h]
\centering

\resizebox{\columnwidth}{!}{
\begin{tabular}{c|c|c|c}
\hline
\multicolumn{1}{l}{Sentence Length Range} & \multicolumn{1}{l}{{\#} of Sentences} & \multicolumn{1}{l}{UAS} & \multicolumn{1}{l}{LAS} \\ \hline\hline
1-10           & 270& 97.24& 96.30\\
11-20          & 764& 97.54& 96.35\\
21-30          & 778& 96.87& 95.83\\
31-40          & 433& 96.73& 95.69\\
41-50          & 135& 97.19& 96.12\\
51-60          & 28 & 94.89& 93.77\\
61-70          & 8  & 94.65& 94.26\\ \hline
all            & 2416               & 96.95& 95.88\\ \hline
\end{tabular}}
\caption{A Table of statistics and performance according to sentence length (based on word count)}
\label{tab:length}
\end{table}
\begin{table}[h]
\centering

\resizebox{\columnwidth}{!}{
\begin{tabular}{c|c|c|c}
\hline
\multicolumn{1}{l}{\;\;Index range\;\;} & \multicolumn{1}{l}{\;\;{\#} of Indices\;\;} & \multicolumn{1}{l}{\;\;UAS\;\;} & \multicolumn{1}{l}{\;\;LAS\;\;} \\ \hline\hline
1-10    & 23269     & 97.77     & 96.84     \\
11-20   & 18283    & 96.39     & 95.14     \\
21-30   & 10124     & 96.32     & 95.22     \\
31-40   & 3829     & 96.63     & 95.59     \\
41-50   & 966     & 96.07     & 95.13     \\
51-60   & 188     & 95.74      & 95.21     \\
61-70   & 25       & 92.00     & 88.00     \\ \hline
all & 56684    & 96.95     & 95.88     \\ \hline
\end{tabular}}
\caption{A Table of statistics and performance according to index range}
\label{tab:length2}

\vskip -.14in
\end{table}

\subsection{Decoder-based Models}
\label{sec:decoder}
The proposed \textbf{SPT-DP} method entirely operates at the \textbf{text level} and it enables our model to have both of the easily applicable and trainable abilities. In this section, we validate its feasibility on different language model architectures, including encoder-decoder models and decoder-only models. In these models, their decoders generate dependency parsing results of $P_{ref}$ and $P_{\mathcal{L}}$ along with other input sequence. As you can see in Table \ref{tab:ptb_decoders_results}, decoder-based models achieve lower performance than encoder-only models. One possible reason for this lower performance is that this decoding method relies on previously generated tokens and it can lead to error propagation; mistakes in earlier tokens are accumulated and it can degrade the accuracy of subsequent predictions.

\begin{table}[h]
\centering
\resizebox{\columnwidth}{!}{
\begin{tabular}{@{}lcc@{}}\hline

Decoder-based Model \; \; \; \; \; \; \; \; \; \;   & UAS & LAS  \\ \hline
\hline
XLNet-large(encoder) &96.95 &95.88\\ \hline
T5-base              & 95.30  & 93.86 \\
Bart-large              & 95.84  & 94.73 \\
Llama3.2-3B              & 94.55  & 93.27 \\
Qwen2.5-3B                & 94.97  & 93.81  \\
\hline
\end{tabular}}
\caption{Results on PTB with Decoder-based models}
\label{tab:ptb_decoders_results}
\vskip -.12in
\end{table}

\subsection{Ablation Study for Prompts}
First, we aim to examine the impact of each proposed prompt on parsing performance. We conduct additional experiments by excluding each of $P_{abs}$ and $P_{{POS}}$ to verify how important they are for dependency parsing. 
As shown in the Table \ref{tab:prompt result}, the role of POS information is not critical but $P_{abs}$ has a significant impact on performance. Intuitively, when a physically referable index prompt $P_{abs}$ exists in the input, the model can effectively refer it through transformer's attention mechanism. A detailed analysis is provided in Appendix \ref{sec:attention}.

\begin{table}
\centering
\resizebox{1\columnwidth}{!}{
\begin{tabular}{@{}lcc@{}}\hline
Method  & UAS & LAS \\ \hline
\hline
SPT-DP                       & 96.95  & 95.88 \\
SPT-DP (w/o $P_{abs}$)       & 94.28 \textcolor{blue}{(-2.67)}  & 92.63\textcolor{blue}{ (-3.25)} \\ 
SPT-DP (w/o $P_{POS}$)       & 96.76\textcolor{blue}{ (-0.19)}  & 95.66\textcolor{blue}{ (-0.22)} \\
\hline
\end{tabular}}
\caption{Effect of prompts on PTB dataset.}
\label{tab:prompt result}
\vskip -.12in
\end{table}

\subsection{Inference Efficiency}
\label{sec:efficiency}
Hexatagging \cite{amini-etal-2023-hexatagging}, which is utilizing sequential labeling, achieves a processing speed 10 times faster than the biaffine model because it does not require additional modules, and similarly to our model, DPSG \cite{lin2022dependency} also adopts a text-to-text approach on generative model.

As a result, although our approach has to increase sentence length due to added prompts, it obtains faster inference speed than other conventional methods because it relies solely on a pre-trained model. 

\begin{table}[h]
\centering
\resizebox{\columnwidth}{!}{
\begin{tabular}{@{}l|ccc@{}}\hline
\multirow{2}{*}{Dataset}& \multicolumn{3}{c}{Speed(sent/s)}\\
  & SPT-DP & Hexatagging & DPSG \\ \hline\hline
 PTB-test  & 39.77 & 28.42  & - \\ 
 UD 2.2 (bg-dev) & 38.58 & -& 0.85 \\ \hline
\end{tabular}}
\caption{Processing speed (sentences per second) of different models on PTB (test) and UD 2.2 (bg-dev)}
\label{tab:speed}
\vskip -.12in
\end{table}

\section{Conclusions}

In this paper, we introduce \textbf{SPT-DP}, as a structuralized prompt template-based dependency parsing method. Our approach enables text-to-text dependency parsing through prompt engineering by utilizing additional tokens while relying solely on pre-trained encoder models without requiring any additional modules. Despite relying solely on a pre-trained encoder model, our proposed method achieves performance comparable to existing models. Through experiments on the UD 2.2, we integrated dependency relation labels to develop a universal model applicable across 12 languages. This model not only enables multi-language dependency parsing within a single model but also demonstrates the ability to generalize to unseen languages to some extent. Finally, we applied our method to decoder-based models, demonstrating its applicability across different model types. Therefore, our method has several strong points; it can be easily applied to various pre-trained models appropriate for the target language or training environments, 
and it achieves fast inference speeds.

\section*{Limitations}
In our method, there is a limitation with sequence length. Although sentences with too many words are occurred in rare cases, additional prompts also increase linearly with the number of words, which can make it difficult to use for encoder models with a short maximum length. In addition, additional research is needed to perform semantic dependency parsing with a dynamic number of relationships.

\section*{Ethics Statement}
We perform dependency parsing using a pre-trained model. The datasets may contain ethical issues or biased sentences, but the model does not influence them through dependency parsing.

\bibliography{custom}

\appendix
\begin{table*}[h]
\resizebox{\textwidth}{!}{
\begin{tabular}{llll}
\hline
Notations & Components & Type & Description \\
\hline
\hline
$w_i$& - & String & Text representation of i'th word in sentence \\
$S$ & $w_i$ & String& Sentence \\
$H_i$& - & Integer& Index of the head word\\
$L_{(w_i,w_{H_i})}$& - &String & Dependency relation label between $w_i$$w_{H_i}$ \\
$r_i$& $H_i,L_{(w_i,w_{H_i})}$& Set & Dependency relation of $w_i$\\
$R_S$& $r_i$ & Set & Dependency relations of sentence S \\
$\mathcal{L}$& $L_{(w_i,w_{H_i})}$ & Set & Predefined dependency relation labels\\
$S_{dep}$& $w_i,r_i$ & Set & Sentence that include dependency parsing information\\
$P_{abs}$ & "[$1$]","[$2$]",\ldots,"[$n$]" & Set of strings & absolute index prompts that located at the first position of the prompt set. \\
$P_{ref}$ & "[$H_1$]","[$H_2$]",\ldots,"[$H_n$]" & Set of strings & reference index prompts that located in the second position of the prompt set.\\
$P_{idx}$ & \{$P_{abs}$, $P_{ref}$\} & Set of strings & String form of index prompts, indies are encapsulated with "[]" \\
$P_{\mathcal{L}}$ & "[acomp]", "[advcl]", \ldots , "[xcomp]" & Set of strings & Prompt tokens of Dependency relation labels , string of labels are encapsulated with "[]"\\
$P_{POS}$ & "[NN]", "[NNP]", \ldots , "[WRB]" & Set of strings & Prompt tokens of pos-tags $\mathcal{P}$, string of pos-tags are encapsulated with "[]"\\
$P_{MASK}$ & "[HEAD]","[DEP]" & Set of strings & Masking prompts for training\\
$D$ &$P_{idx}$,$P_{\mathcal{L}}$,$P_{POS}$,$S$ & String & Structuralized Prompt Template\\ 
$D_M$ & $P_{MASK}$,$P_{POS}$,$S$ & String & Masked Structuralized Prompt Template \\
\hline
\end{tabular}}
\caption{Notations}
\label{tab:notation}
\end{table*}

\section{Notations}

\section{Implementation Details}

For experiments for PTB, xlnet-large-cased\footnote{\url{https://huggingface.co/xlnet-large-cased}} are used. For experiments for UD 2.2, bert-multilingual-cased\footnote{\url{https://huggingface.co/bert-base-multilingual-cased}}, xlm-roberta-large\footnote{\url{https://huggingface.co/FacebookAI/xlm-roberta-large}} and xlnet-large-cased are used. For decoder-based models, T5-base\footnote{\url{https://huggingface.co/google-t5/t5-base}}, Bart-large\footnote{\url{https://huggingface.co/facebook/bart-large}}, Qwen2.5-3B\footnote{\url{https://huggingface.co/Qwen/Qwen2.5-3B}}, Llama3.2-3B\footnote{\url{https://huggingface.co/meta-llama/Llama-3.2-3B}} are used. For the Korean Sejong dataset, RoBERTa-large\footnote{\url{https://huggingface.co/klue/roberta-large}}, a pre-trained model for the Korean language, is used. Experiments are conducted on an NVIDIA RTX A6000. The models are fine-tuned with a batch size of 8, a learning rate of 1e-5, and 10 training epochs. Training is performed using the linear scheduler and AdamW optimizer.



\section{Efficiency test}
For the efficiency test between Hexatagging and SPT-DP, we used the PTB test dataset to evaluate the speed of dependency parsing. For the efficiency test between DPSG and SPT-DP, we used the UD 2.2-bg dev dataset to evaluate the speed of dependency parsing. We set the batch size to 1 and conducted the experiment under the same conditions using a single A6000 GPU.

\section{Decoder-based Model}
Unlike encoders, the decoder-based model required constrained generation. During the inference stage, contents other than $P_{ref}$ and $P_{\mathcal{L}}$ were forcibly inserted into the sequence at intervals, allowing the model to perform accurate dependency parsing in a restricted environment.

\section{Analysis : Ablation Study for Prompts}
\label{sec:attention}
\begin{figure}[h]
 \includegraphics[width=\columnwidth]{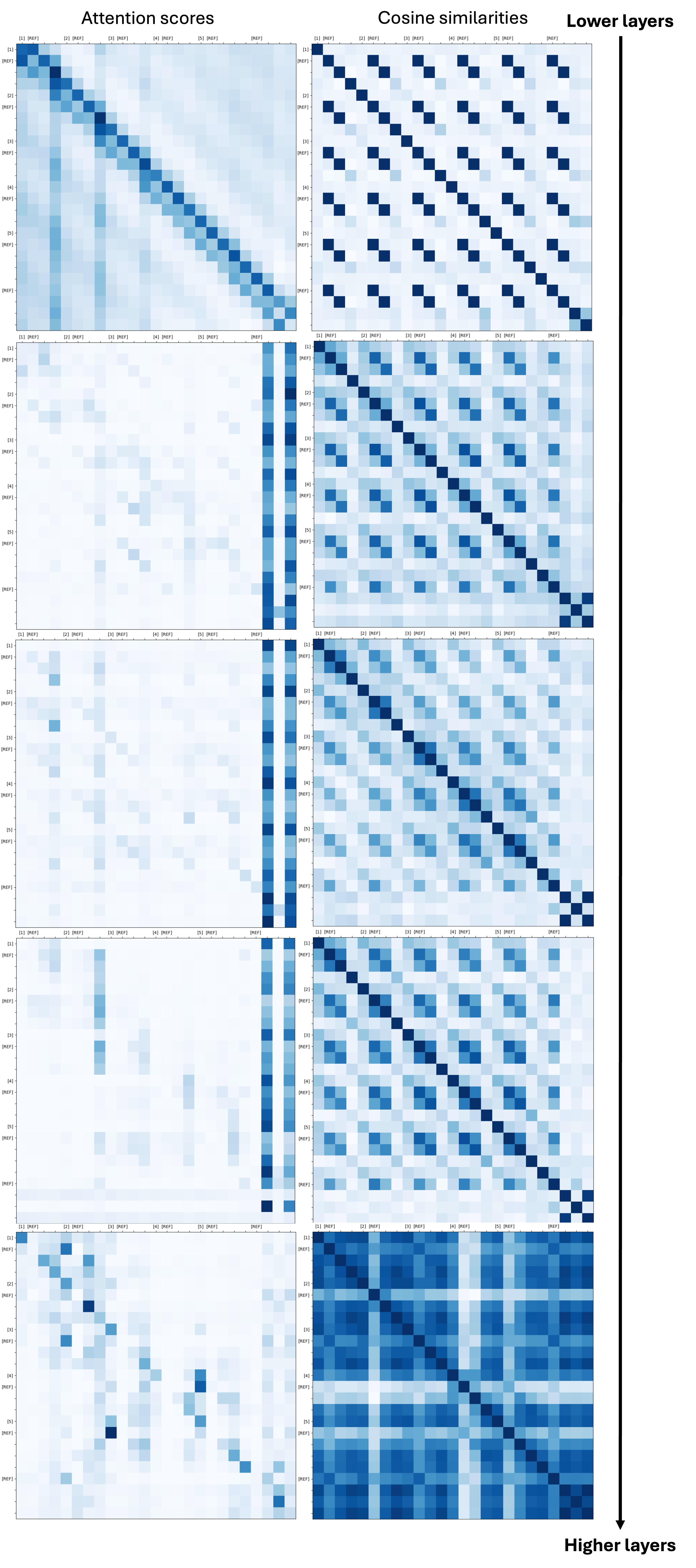}
 \caption{A heatmap of Attention scores and cosine similarities in hidden layer}
 \label{fig:heatmap}
\end{figure}

To elaborate further, in experiments without $P_{abs}$, the order of the template implicitly replaced $P_{abs}$ and was used for prediction.
The experimental results (Table \ref{tab:prompt result}) indicate that the explicit presence of $P_{abs}$ (physically exists), allowing for direct reference, plays a crucial role in dependency parsing through attention.
In Figure \ref{fig:heatmap}, presents a heatmap representation of attention scores across layers and the cosine similarity of each hidden state. The attention scores on the left show that as the layers progress, the values converge, with the token in $P_{ref}$ exhibiting high attention scores toward $P_{abs}$ , which it is supposed to reference.

\section{Licenses}
The PTB dataset is licensed under LDC User Agreement. The UD 2.2 dataset is licensed under the Universal Dependencies License Agreement.

\end{document}